\documentclass[10pt,twocolumn,letterpaper]{article}

\usepackage[pagenumbers]{cvpr} 

\usepackage{xspace}
\usepackage{layouts}
\usepackage{adjustbox}  
\usepackage{multirow}  
\usepackage{makecell}  
\usepackage{amsmath}
\usepackage{threeparttable}  
\usepackage[ruled,linesnumbered]{algorithm2e}
\usepackage{bm}  
\usepackage{enumitem}  
\usepackage{diagbox}  
\usepackage{caption} 
\usepackage{tabularx} 
\usepackage{wrapfig}  

\newcommand{\tref}[1]{Tab.~\ref{#1}}

\newcommand{\eref}[1]{Eq.~(\ref{#1})}

\newcommand{\fref}[1]{Fig.~\ref{#1}}

\newcommand{\sref}[1]{Sec.~\ref{#1}}

\newcommand{\V}[1]{\ensuremath{\bm{#1}}}

\definecolor{cvprblue}{rgb}{0.21,0.49,0.74}
\usepackage[pagebackref,breaklinks,colorlinks,allcolors=cvprblue]{hyperref}

\def\paperID{*****} 
\def\confName{CVPR}
\def\confYear{2026}

\title{PacTure: Efficient PBR Texture Generation on Packed Views with Visual Autoregressive Models}

\author{
Fan Fei$^{1,2,4,5}$~~~Jiajun Tang$^{1,2}$~~~Fei-Peng Tian$^{4}$~~~Boxin Shi$^{1,2,5\dagger}$~~~Ping Tan$^{3,4}$
\\{\small $^1$ State Key Laboratory of Multimedia Information Processing, School of Computer Science, Peking University}
\\{\small $^2$ National Engineering Research Center of Visual Technology, School of Computer Science, Peking University}
\\{\small $^3$ The Hong Kong University of Science and Technology}~~~~~~{\small $^4$ Light Illusions}
\\{\small $^5$ PKU-AI$^2$ Robotics Joint Lab of Embodied AI}
\\\small\texttt{\{fanfei,~jiajun.tang,~shiboxin\}@pku.edu.cn}
\\\small\texttt{flybirdtian@gmail.com~~pingtan@ust.hk}
}

\begin{document}
\maketitle

\begin{abstract}
We present PacTure, a novel framework for generating physically-based rendering (PBR) material textures for an untextured 3D mesh from a text description.
Existing 2D generation-based texturing approaches either generate textures sequentially from different views, resulting in long inference times and globally inconsistent textures, or adopt multi-view generation with cross-view attention to enhance global consistency, which, however, limits the resolution for each view.
In response to these weaknesses, we first introduce view packing, a novel technique that significantly increases the effective resolution for each view during multi-view generation, without imposing additional inference cost.
Unlike UV mapping, it preserves the spatial proximity essential for image generation and maintains full compatibility with current 2D generative models.
To further reduce the inferencing cost, we enable fine-grained control and multi-domain generation within the next-scale prediction autoregressive framework, creating an efficient multi-view PBR generation backbone.
Extensive experiments show that PacTure outperforms state-of-the-art methods in both quality and efficiency.
\end{abstract}

\section{Introduction}

\begin{figure*}[!t]
    \centering
    \includegraphics[width=\textwidth]{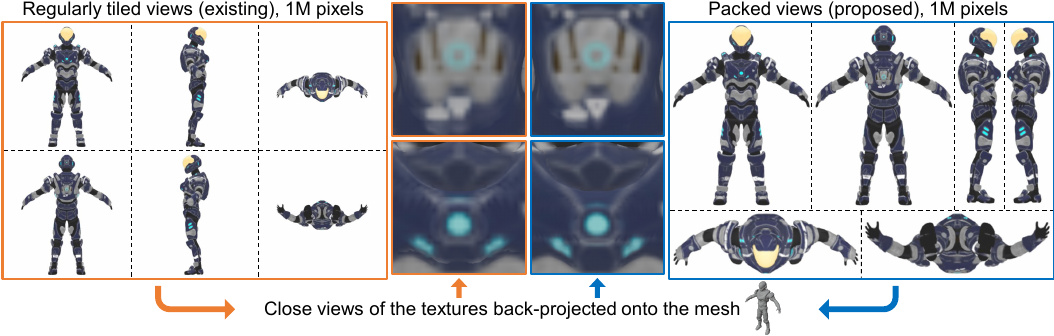}
    \caption{To generate texture with image generation models, we use a packed view representation (as both conditions and targets; we show only targets here) that compactly packs multi-view maps onto an atlas.
    In this way, the effective resolution for each view is significantly increased without incurring additional generation costs.
    }
    \label{fig: teaser}
\end{figure*}
Physically-based rendering (PBR)~\cite{Microfacet07} textures are an important part of 3D asset, as they provide photorealistic view-dependent appearance under varying lighting conditions, and so are preferred in industries such as gaming, movies, virtual/augmented reality (VR/AR), and simulation for embodied AI, as shaded texture maps cannot be easily relit.
Current texture generation (\ie, texturing) approaches~\cite{TEXTure23, Text2Tex23, Paint3D24, TexFusion23, Hunyuan3D2.025, GenesisTex24, InTex24, Meta3DTextureGen24, Meta3DAssetGen24, VCDTexture24, CLAY24, ECCVTexGen24, MakeATexture25, ColControl24, MVPaint25, MCMat24, TexPainter24, Pandora3D25, SyncMVD24, PointUV23, TEXGen24, UV3Ted24, TexOct24, TexGaussian25} seek to leverage powerful generative priors to automatically synthesize high-quality PBR textures directly from input geometry and semantic guides such as text prompts, thereby accelerating 3D content creation and reducing artist workload.

There are generally two types of texturing paradigms that require no inference-time optimization: native generation in 3D space~\cite{PointUV23, TEXGen24, UV3Ted24, TexOct24, TexGaussian25, SAR3D25}, and projection of generated 2D images onto mesh surfaces~\cite{TEXTure23, Text2Tex23, Paint3D24, TexFusion23, Hunyuan3D2.025, GenesisTex24, InTex24, Meta3DTextureGen24, Meta3DAssetGen24, VCDTexture24, CLAY24, ECCVTexGen24, MakeATexture25, ColControl24, MVPaint25, MCMat24, TexPainter24, Pandora3D25, SyncMVD24}.
The latter typically involves rendering maps of geometric properties (such as depth) from different viewpoints, then passing them as conditions to a generative model for pixel-aligned texture synthesis, followed by back-projecting the generated textures onto the model surface, often parametrized as a UV map.
This approach is particularly attractive due to its ability to harness powerful 2D generative models~\cite{SD22, SD324, SDXL24, Emu23, Diffusion43D25, TamingDiffusion24} pretrained on extremely large-scale 2D image datasets, which results in better generalization ability and lower fine-tuning cost.

However, there are two major weaknesses in such 2D-based pipelines: (i) \emph{low effective resolution} and (ii) \emph{long inferencing times}.
Early works of this kind~\cite{TEXTure23, Text2Tex23} employed a sequential view sampling strategy and iteratively generated textures visible from one viewpoint at a time.
While this approach requires no further fine-tuning of the pretrained generative models, it has long inferencing times, and also leads to globally incoherent textures, commonly referred to as the Janus problem~\cite{Meta3DTextureGen24}.
To enhance global multi-view consistency, subsequent methods concatenated multi-view maps channel-wise and introduced cross-view attention~\cite{ColControl24, GenesisTex24, MCMat24, 3DEnhancer25} or concatenated them spatially into a grid~\cite{CLAY24, Meta3DAssetGen24, FlashTex24, Pandora3D25} to facilitate communication between views.
However, using cross-view attention significantly increases memory and time costs, thus limiting the resolution allocated for each view.

We introduce a novel technique we call \emph{view packing} to \emph{increase the effective resolution for each view} without increasing the resolution to be generated, or inference costs.
We concatenate the multi-view maps spatially like other methods, but we note that naively tiling square maps into a regular grid (see \fref{fig: teaser}(left)) can be inefficient.
The aspect ratios of the bounding boxes of visible regions can vary greatly between views; as a result, squeezing each into a square leaves extensive background areas that do not contribute to the final texture upon back-projection.
Moreover, uniformly allocating the same area for each view disregards the actual surface area seen from each viewpoint, thus potentially wasting the limited generative power and computational resources in small regions and overlooking large regions that require detail.
Our view packing (see \fref{fig: teaser}(right)), inspired by the chart packing process in UV unwrapping~\cite{AtlasGen02}, instead maximizes the area occupied by foreground pixels in multi-view grids to increase the effective resolution and enhance texture details (see \fref{fig: teaser} (middle)), while preserving the spatial proximity, and so remaining fully compatible with existing 2D generative models.
It formulates the arrangement of multi-view maps onto the atlas as a 2D rectangle bin packing problem~\cite{binpacking87, binpacking10} and solves it to determine the optimal enlargement of views that minimizes wasted space.
Compared to conventional view grid tiling, this adaptive view packing strategy substantially increases the effective resolution of each view, raising the foreground pixel ratio by $1.87\times$ on average on our dataset, consequently improving the quality of the back-projected textures.

To \emph{further reduce the inferencing cost} of PBR texture generation, we turn to visual autoregressive (VAR) models~\cite{VAR24, Infinity25}, which are also known as next-scale prediction models and have recently demonstrated superior efficiency, quality, and scalability for text-to-image (T2I) synthesis compared to diffusion-based models~\cite{SD324, SDXL24, PIXARTSigma24} and next-token prediction models~\cite{LLamagen24, HART24, Emu23} that prior texturing works relied upon.
However, it is non-trivial to adopt VAR T2I models for PBR texturing, as it demands geometric faithfulness, multi-view consistency, and multi-domain outputs (diffuse albedo, metallicity, and roughness).
To address these challenges, we introduce several efficient adaptations within the VAR framework to enable fine-grained control and multi-domain output, as well as a two-stage single-to-multi-view generation framework to improve quality and flexibility.
Combined with our view packing technique, they provide our \emph{PacTure} method, which achieves state-of-the-art PBR texturing quality and efficient training ($\approx 240$ GPU hours on A6000 GPUs) and inferencing ($\approx 30$ s on a single A6000 GPU for each mesh).
In summary, our main contributions are as follows:
\begin{itemize}
    \item a novel view packing technique that substantially improves effective resolution with no extra inferencing cost for 2D generation-based texturing methods,
    \item PacTure, a PBR texturing pipeline that efficiently enables fine-grained control and multi-domain output within the VAR T2I model, providing fast training and inferencing.
\end{itemize}
Extensive experiments demonstrate that PacTure outperforms state-of-the-art methods in both the quality of the generated PBR textures and inferencing time.

\section{Related Work}

\subsection{Prompt-based Texture Generation}

Given an untextured 3D mesh and an input prompt (which may be a text description, reference image, or both), the objective of texture generation is to assign colors or PBR material properties to the mesh surface so that rendering the mesh yields images that are semantically faithful to the provided prompt.
Early approaches~\cite{Text2Mesh22, Texturify22, GET3D22, CLIPMesh22, LatentNeRF23, Paintit24, TextureDreamer24, DreamMat24} adopted a test-time optimization pipeline that leverages differentiable rendering with guidance from CLIP~\cite{CLIP21}, GANs~\cite{GAN14}, or score distillation sampling (SDS)~\cite{DreamFusion23}.
These methods are computationally intensive, typically requiring tens of minutes to several hours to texture a single mesh.
To address this inefficiency, more recent methods employ image generation models~\cite{SD22, Emu23, T2IBDMT2025} conditioned on geometric information (\eg, depth~\cite{ControlNet23}), and directly back-project the images, sampled from certain viewpoints, onto the mesh as textures.
This strategy dramatically accelerates inferencing as it minimizes the need for optimization.
Within this paradigm, sequential-view methods~\cite{TEXTure23, Text2Tex23, InTex24, Paint3D24, ECCVTexGen24} sequentially paint the mesh one view at a time.
These approaches are thus sensitive to the selection and order of viewpoints, and are susceptible to the Janus problem, in which the 2D generative model, limited to a single viewpoint at a time, cannot enforce global consistency across distant views.
In contrast, concurrent-view methods advance by generating images from multiple viewpoints simultaneously, either by arranging views in grids~\cite{CLAY24, Meta3DAssetGen24, Meta3DTextureGen24, MatAtlas24, FlashTex24, MVPaint25, Pandora3D25} or by incorporating cross-view attention mechanisms~\cite{ColControl24, GenesisTex24, MCMat24}.
These techniques enhance global consistency and effectively mitigate the Janus problem, but also increase the memory and time required for generation.
Notably, some concurrent-view methods only synchronize between neighboring views in UV space~\cite{TexFusion23, SyncMVD24, GenesisTex24, TexPainter24} without enabling long-range information exchange.
There are also methods that directly generate in UV space, where surfaces are cut, flattened, and packed onto an atlas.
However, as these steps disrupt the spatial continuity essential for 2D generation and make UV space unsuitable for image generation models, works exploiting UV spaces only use them in post-processing~\cite{Paint3D24, Meta3DTextureGen24, Meta3DAssetGen24}, by incorporating additional 3D cues~\cite{TEXGen24, PointUV23}, or are limited to generation in UV spaces with regular topology~\cite{TexDreamer24}.
Our PacTure falls within the concurrent-view category as it generates texture images from six viewpoints simultaneously.
Distinctively, we pack the views into a grid rather than simply tiling them, thereby increasing effective resolutions and preserving spatial adjacency within each view.
There are also methods that do not heavily rely on a pretrained 2D generative model, such as part-based material library retrieval methods~\cite{MakeItReal24, MatAtlas24, MaterialSeg3D24, MaPa24} and 3D native texture generation methods~\cite{PointUV23, TEXGen24, UV3Ted24, TexOct24, TexGaussian25}.
However, the former struggle to generate semantically meaningful textures within a part, and the latter have a huge demand for high-quality 3D data and computational resources for training and suffer from suboptimal generalization.

\begin{figure*}[t!]
    \centering
    \includegraphics[width=\textwidth]{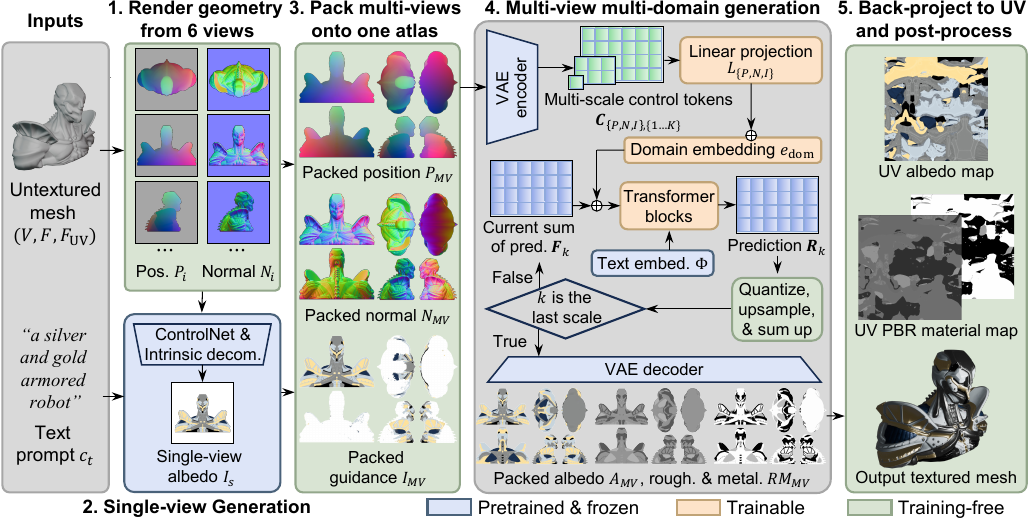}
    \caption{
    Overview of our pipeline, with the following steps.
    1) Given the input untextured mesh and text description, we first render geometry condition maps from 6 views (see \sref{subsec: pre-processing}).
    2) An off-the-shelf generative model with intrinsic decomposition is used to generate a single-view albedo map (see \sref{subsec: two stage}).
    3) The geometry conditions and the single-view albedo map are packed compactly onto the atlas as the control images for the subsequent multi-view generation (see \sref{subsec: view packing}).
    4) We adjust the architecture of the visual autoregressive model to enable multi-view multi-domain generation (see \sref{subsec: adapting VAR}).
    5) The generated multi-view PBR material maps are back-projected and post-processed in $UV$ space (see \sref{subsec: back-projection}), resulting in the textured mesh with $UV$ PBR material maps.
    }
    \label{fig: pipeline}
\end{figure*}

\subsection{Generative Models with Pixel-aligned Control}

For texture generation tasks, it is essential that the generated texture strictly conforms to the given geometry: otherwise, the texture appears unrealistic when back-projected onto the mesh surface.
Accordingly, to ensure geometric fidelity, many texture generation approaches integrate spatially aligned control images, such as depth maps, into the image synthesis process.
For diffusion-based models, ControlNet~\cite{ControlNet23} uses a trainable copy of the diffusion U-Net encoder~\cite{UNet15} to encode the control image, adding its features to those of the original network.
Other methods propose lighter-weight adapters~\cite{T2IAdapter24, ControlNeXt24, OminiControl24}, unification of various control signals~\cite{UniControlNet23, UniControl23, OminiControl24} for improved parameter efficiency and reduced training costs, or the incorporation of cycle consistency to reinforce conditional control~\cite{ControlNetPP24}.
For autoregressive models, ControlAR~\cite{ControlAR24} employs a lightweight pretrained vision transformer (ViT), DINOv2-S~\cite{DINOv224}, to encode the control image into a token sequence, which is then added token-wise to the image token sequence.
PIXART-$\delta$~\cite{PIXARTDelta24} adapts the ControlNet~\cite{ControlNet23} paradigm to the ViT architecture used in PIXART-$\alpha$~\cite{PIXARTAlpha24} by copying a different set of blocks.
ControlVAR~\cite{ControlVAR24} concurrently generates image and control outputs, enabling multiple tasks such as control-to-image generation via inference-time teacher forcing.
CAR~\cite{CAR24} injects multi-scale control features into the image token map through an additional transformer.
Our method adapts Infinity~\cite{Infinity25}, a next-scale prediction-based autoregressive model, as the backbone for multi-view generation.
Specifically, we utilize its VQVAE~\cite{VQVAE17} encoder to map the control image onto multi-scale residual token maps, which are linearly projected and subsequently added to the image token maps at each scale to facilitate fine-grained, spatially aligned control.
Compared to prior works, our method requires a negligible number of additional parameters~\cite{ControlNet23, PIXARTDelta24, CAR24}, imposes no further computation or memory overhead (unlike~\cite{ControlVAR24, OminiControl24} which require extra attention or token map generation), and eliminates the need for alignment between feature spaces of different pretrained networks~\cite{ControlAR24}.
Consequently, our approach achieves fine-grained control with rapid training convergence and low inference latency.

\section{Method}

Given a UV-unwrapped surfaced mesh $(V, F, F_{\rm UV})$, with vertices $V$, triangular faces $F$, and $F_{\rm UV}$ denoting face coordinates in UV space, together with a text prompt $t$, our PBR texture generation pipeline PacTure aims to generate a UV-space albedo map $A_{\rm UV}$ and a roughness-metallicity (RM) map $R_{\rm UV}$ for the mesh that gives appearances faithful to the prompt.
\fref{fig: pipeline} illustrates the pipeline of PacTure, which is further explained in the following sections; the appendix gives additional implementation details.

\subsection{Rendering Multi-view Geometry Conditions}
\label{subsec: pre-processing}

Firstly, given the input untextured mesh, we render multi-view geometric condition maps for it from six canonical viewpoints (front, rear, left, right, top, and bottom), serving as the conditions for the subsequent multi-view generative model.
We pre-process the input mesh and render the following maps: the position map $P_{i}$, world-space surface normal map $N_{i}$, depth map $D_{i}$, and alpha map $M_{i}$ from each viewpoint $i \in \{ 1, \dots, 6 \}$:
\begin{align}
\label{equ: render geo maps}
P_i, N_i, D_i, M_i = \operatorname{Render}(V, F, {\rm View}_i) ,\quad \forall i.&&
\raisetag{11.5pt}
\end{align}
We utilize position and world-space normal maps because their view-invariance (which does not hold for depth maps) facilitates the capture of cross-view correspondences by the generative model.
However, we still retain the depth map for subsequent single-view generation using models that operate on depths.
The normal maps also supply high-frequency geometric details that are less discernible in position maps.

\subsection{Single-view Generation as Guidance}
\label{subsec: two stage}
A straightforward approach to multi-view material map generation would be to directly synthesize the maps from the given geometry condition and prompts.
However, we find that a two-stage process, inspired by recent advances in texturing~\cite{Pandora3D25}, comprising single-view generation followed by multi-view generation, is beneficial.
Thus, our second step utilizes an off-the-shelf image generation model $\mathcal{G}_{\rm SV}$ and an intrinsic decomposition model $\operatorname{IntrinsicDecompose}$ to generate a single-view diffuse albedo image of view $s$ as guidance.
Given the depth map $D_s$ and the prompt $c_t$, the single-view albedo image $I_s$ is given by:
\begin{align}
\label{equ: stage1}
    I_s = \operatorname{IntrinsicDecompose}(\mathcal{G}_{\rm SV}(D_s, c_t)).
\end{align}
$I_s$ is then spatially aligned with geometry maps from view $s$ during view packing (see \sref{subsec: view packing}), and functions similarly to the pixel-aligned guidance of the multi-view generation process.

Introducing single-view guidance has two advantages.
Firstly, it produces higher-quality results: single-view models benefit from a more focused learning objective, being trained solely on large-scale curated 2D image datasets, which typically consist of more diverse and aesthetically pleasing images than multi-view datasets that are often restricted to renderings from 3D datasets like Objaverse~\cite{Objaverse23}.
Incorporating single-view guidance leads to multi-view maps with richer details.
Secondly, this approach enhances flexibility and user control.
It readily accommodates replaceable modules in the single-view model (\eg, IP-Adapter~\cite{IPAdapter23} for prompting with an input image), eliminating the need for complex customizations within the multi-view generation backbone.
Moreover, users can easily select a preferred image or replace the first-stage output with any custom image as needed prior to multi-view generation.

\subsection{View Packing}
\label{subsec: view packing}

We next propose a novel view packing technique to pack condition maps from different views into a single atlas to maximize pixel utilization, yielding the multi-view condition maps $P_{\rm MV}, N_{\rm MV}, I_{\rm MV}$ for the subsequent multi-view generation process:
\begin{align}
    P_{\rm MV}, N_{\rm MV}, I_{\rm MV} = \operatorname{Pack}(P_*, N_*, M_*, I_s),
\end{align}
where a subscript $*$ ranges over all views.
This technique formulates the arrangement of multi-view maps into the atlas as a 2D rectangle bin packing problem where the rectangles are the bounding boxes of the alpha maps.
It employs a coarse-to-fine binary search strategy to determine the maximal enlargement ratios for all views that enable them together to fit within the atlas.
In this way, the generative model's resource wastage due to blank pixels or disproportionate area allocation is minimized, and visual legibility, crucial for 2D generation, is fully retained.

Notably, the multi-view version $I_{\rm MV}$ of the single-view guidance $I_s$ is obtained by first setting guidance maps for other views as blank and copying pixels in view $s$ into other views according to positions and normals to enhance the guidance signal, as shown in \fref{fig: pipeline}(bottom left).

\begin{figure}[!t]
    \centering
    \includegraphics[width=\columnwidth]{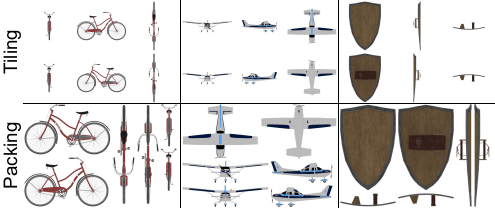}
    \caption{Comparison of our proposed view packing and the commonly-used regular grid view tiling.
    }
    \label{fig: view packing}
\end{figure}

The packing process proceeds as follows.
Given a target pixel budget (\eg, 1M), we set the size of the atlas, into which the geometry condition maps must be compactly packed, using an aspect ratio of 3:2 (\eg, $1248 \times 832$), as there are 6 views.
In the preparatory stage, we compute the bounding box of the alpha map $M_i$ for each of the six views, which reflects the visible region from each viewpoint, and use the dimensions of these bounding boxes as the base sizes of rectangles to be packed.
We then employ binary search to find a global enlargement ratio such that the expanded rectangles fit into the atlas, using the MaxRects bin packing algorithm for a feasibility check.
The algorithm receives the height and width of each rectangle (\ie, the enlarged views) to pack and also the height and width of the bin (\ie, the atlas) into which the rectangles are packed and returns the minimum number of bins such that the rectangles can be packed into these bins.
We deem the packing to be successful if the number of required bins is one.
After fixing the global enlargement ratio, we further attempt paired enlargement of opposite views iteratively from larger pairs to smaller pairs, using a similar process to the above.

The above carefully designed global-to-local, large-to-small strategy effectively maximizes the foreground pixel ratio while maintaining cross-view relative sizes, ultimately achieving a balanced and efficient utilization of the generative model's capacity.
\fref{fig: view packing} visually compares the multi-view images resulting from our view packing technique and from the conventional regular tiling strategy (\ie, one view per slot).
For our dataset, our view packing strategy increases the average foreground pixel proportion in the packed multi-view image from 22.6\% to 42.2\%, a $1.87\times$ improvement.
The total covered area of the padded bounding boxes also averages 88.3\%, indicating that our strategy yields a near-optimal packing efficiency, for bounding boxes that do not overlap.

\subsection{VAR Models for Multi-view PBR Generation}
\label{subsec: adapting VAR}

We construct our multi-view generative backbone $\mathcal{G_{\rm MV}}$ upon Infinity, a recent T2I generative model based on the VAR model; we introduce structural adaptations to enable fine-grained control and multi-domain output.
The backbone generates the multi-view albedo $A_{\rm MV}$ and RM map $R_{\rm MV}$ that follow the text prompt $c_t$.
They are pixel-aligned with the condition maps $P_{\rm MV}, N_{\rm MV}, I_{\rm MV}$:
\begin{align}
\label{equ: stage2}
    A_{\rm MV}, R_{\rm MV} = \mathcal{G}_{\rm MV}(P_{\rm MV}, N_{\rm MV}, I_{\rm MV}, c_t).
\end{align}
We provide a compact reformulation below to provide a self-contained explanation.

\subsubsection{Formulation}
In contrast to next-token prediction~\cite{VQGAN21} or denoising diffusion models~\cite{SD22}, the VAR model adopts a coarse-to-fine next-scale prediction paradigm.
At each inference step $k = 1 \to K$, it generates an entire token map with resolution $h_k \times w_k$, beginning at $1\times 1$ and growing larger until reaching the desired output resolution $h_K \times w_K$, thereby enabling faster inferencing and yielding superior generation quality.
Infinity introduces two major advances over the basic VAR model: (i) an effectively unbounded vocabulary size realized by lookup-free bitwise quantization~\cite{BSQ24}, and (ii) the capacity for T2I generation via integration with a text encoder.

The architecture of Infinity comprises two primary components: a VQVAE acting as the visual (de)tokenizer, including an encoder $\mathcal{E}$, a quantizer $\mathcal{Q}$, and a decoder $\mathcal{D}$, and a decoder-only transformer serving as the generative core.
For inferencing, Infinity employs a residual prediction framework.
Given a text prompt $c_t$, at each step $k$, the transformer predicts the next-scale residual discrete token map $\V{R}_{k} \in \mathbb{V}^{h_{k} \times w_{k}}$ (where $\mathbb{V} = \{-1, +1\}^b$ is the vocabulary of bit tokens with length $b$), conditioned on both the text embedding $\Phi(c_t)$ produced by the Flan-T5 encoder~\cite{FlanT524} and the current sum $\V{F}_{k}$ of previous predictions; $\V{F}_{1}$ is instead the start-of-sequence token:
\begin{flalign}
\label{equ: next-scale prediction}
    &p\left( \V{R}_1, \dots, \V{R}_{K} \right) = \prod_{k=1}^{K} p\left( \V{R}_{k} \,\middle\vert\, \V{F}_{k}, \Phi(c_t) \right), &&
    \\ \V{F}_{k} = & \operatorname{Down}\left(\sum_{i=1}^{k-1}{\operatorname{Up}\left(\mathcal{Q}(\V{R}_i), (h_K, w_K)\right)}, (h_k, w_k) \right),
    && \raisetag{22pt}
\end{flalign}
where $\operatorname{Up/Down}(\V{F}, (h, w))$ up-/down-samples the token map $\V{F}$ to the target resolution $(h, w)$, and the quantizer $\mathcal{Q}$ employs lookup-free binary spherical quantization (BSQ)~\cite{BSQ24}, in which, given a raw feature $z$, the quantized feature $q$ at dimension $j$ is:
\begin{align}
    q_j = \mathcal{Q}(z_j) = \operatorname{sign}(z_j)/\sqrt{b}, \quad \text{for} \; j \in  \{1,\dots,b\}.
\end{align}
The decoder $\mathcal{D}$ gives the final RGB image as
$\widetilde{\V{I}} = \mathcal{D}(\V{F}_{K})$.

For training, the VAE encoder $\mathcal{E}$ transforms a ground-truth (GT) image $\V{I} \in \mathbb{R}^{H \times W \times 3}$ into a raw feature map $\V{F} = \mathcal{E}(\V{I}) \in \mathbb{R}^{h_K \times w_K \times b}$, with $h_K = H/16, w_K=W/16$.
The raw feature map is quantized into multi-scale residual token maps $(\V{R}_1, \dots, \V{R}_{K})$ for teacher-forcing training: at scale $i$, the residual map $\V{R}_i$ serves as the learning target while the cumulative sum serves as the input to the network blocks.

From the above, we can see that Infinity does not inherently support fine-grained control for geometry following or multi-domain output for generating different PBR attributes.
We address these limitations below.

\subsubsection{Fine-grained Control}
Fine-grained control can be decomposed into two key subproblems: (i) encoding the control image, and (ii) guiding the generative process using the encoded control.
For the first subproblem, we freeze the VAE encoder and quantizer, and use them to encode the control image into multi-scale residual token maps.
This strategy ensures that both image tokens and control tokens reside in a unified latent space from the very beginning (though possibly on different manifolds), in contrast to methods~\cite{ControlAR24} that use pretrained feature extractors operating in distinct feature spaces (\eg, DINOv2) or untrained encoders.
Thus, it eliminates the need to introduce additional heavy-weight modules~\cite{ControlNet23, CAR24}, resulting in more rapid training convergence.
For the second subproblem, since control and output images are spatially aligned, we follow common practice~\cite{ControlNeXt24} and directly add the control token maps, each processed by a linear layer trained for each control type, to the image token map prior to the first transformer block at each scale.
Unlike methods that jointly generate both token maps~\cite{ControlVAR24} or incorporate auxiliary transformers for token fusion~\cite{CAR24}, our approach introduces negligible extra computational or memory overhead.

\subsubsection{Multi-domain Output}
To enable dual-domain generation (albedo and roughness-metallicity, where roughness and metallicity are encoded in the green and blue channels, respectively), an intuitive solution would be to expand the input and output channels of the VAE or configure the transformer to generate a double-sized token map, subsequently split into two.
However, fine-tuning the VAE may alter the feature space upon which the pretrained transformer relies, potentially resulting in catastrophic forgetting.
Making the transformer generate two token maps at once also undermines the generation quality, as we later show in \sref{subsec: ablation}.
Therefore, inspired by Wonder3D~\cite{Wonder3D24}, we add domain embeddings to the token map at each scale to inform the transformer of the intended domain for generation.

Let $\{\V{C}_{X, 1},\dots,\V{C}_{X, K}\}$ denote the encoded multi-scale control token maps from the control map $X_{\rm MV}$ ($X \in \{P, N, I\}$), $L_{X}$ be the linear projection for control $X$, and $e_{\rm dom}$ the domain embedding, the right hand side of next-scale prediction (\eref{equ: next-scale prediction}) is modified to:
\begin{align}
    \prod_{k=1}^{K} p_{\rm dom} & ( \V{R}_{k} \,\vert\, \V{F}_{k} + \sum_{X} L_{X}(\V{C}_{X, k}) + e_{\rm dom}, \Phi(c_t) ).
\raisetag{22pt}
\end{align}

\begin{figure*}[t!]
    \centering
    \includegraphics[width=\textwidth]{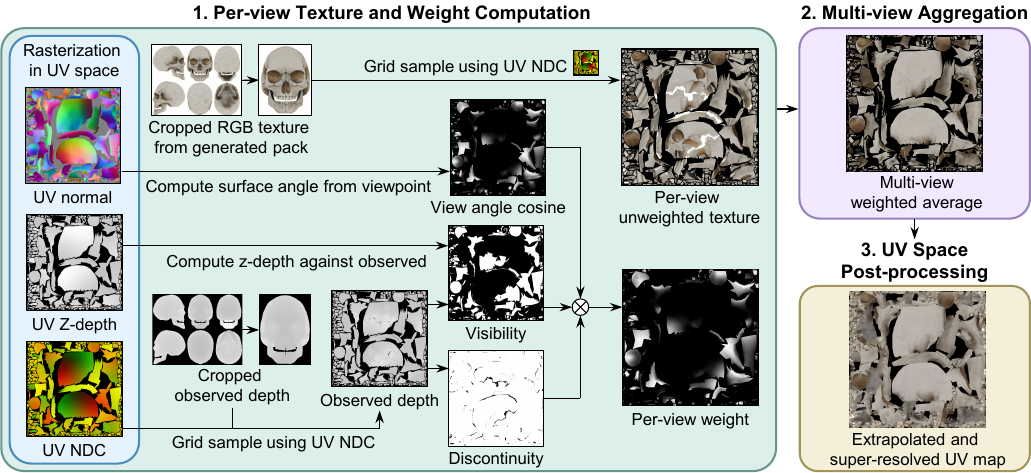}
    \caption{For each PBR attribute (albedo, roughness, metallicity), we merge multi-view texture maps into a UV texture using a direction- and discontinuity-aware back-projection process, using per-view texture and weight computation, multi-view aggregation, and UV space post-processing.}
    \label{fig: back projection}
\end{figure*}

\subsection{Back-projecting Multi-view Textures}
\label{subsec: back-projection}

We next project the multi-view material maps into UV space via view angle- and discontinuity-aware aggregation.
The resulting coarse UV maps are further post-processed using super-resolution and extrapolation-based hole filling to get the final UV material maps $X_{\rm UV}$ ($X \in \{A, R\}$):
\begin{flalign}
    X_{\rm UV} = \operatorname{PostProcess}(\operatorname{BackProject}(X_{\rm MV}, V, F, F_{\rm UV})). &&
\raisetag{11.5pt}
\end{flalign}

Our back-projection process assumes that, in the given UV mapping $F_{\rm UV}$ for the mesh, different triangle faces do not overlap in UV space.
Similar processes have been used in 2D-based texturing methods~\cite{Meta3DTextureGen24, Paint3D24}, and we further adapt them to accommodate packed views.
The process is illustrated in \fref{fig: back projection} and can be separated into three stages: per-view texture and weight computation, multi-view aggregation, and UV space post-processing.

\subsubsection{Per-view Texture and Weight Computation}
During per-view texture and weight computation, we first rasterize the triangles into UV space with respect to each view, rendering normalized device coordinate (NDC) maps (where the first two dimensions represent the location within the view-space image of a surface point), depth maps without considering occlusion (\ie, Z-depth), and camera-space surface normal maps.
The NDC maps are used to index into the generated multi-view textures and the latter two are used to compute aggregation weights for different views.
The NDC maps are modified during view packing to maintain their correctness, so that each one represents locations within the packed map.
Then, again for each view, we sample into the packed multi-view texture map and the depth map considering occlusion (\ie, observed depth) according to the modified NDC map of this view to get the UV texture map, observed depth map, and discontinuity map (by extracting edges in the view-space depth map).
Per-view visibility maps are then obtained by comparing the observed depth against the Z-depth to filter out surfaces that are occluded in a specific view.
To account for the view angle, the view angle map is computed as the cosine between the view angle and the surface normal to the power of 2, after being clipped to $[0,1]$.

\subsubsection{Aggregation and Post-processing}
The view angle and discontinuity-aware weight map for each view is computed as the product of the visibility map, the view angle map, and the discontinuity map, and is used as the weight to get the weighted average of per-view UV texture maps as the back-projected texture map during multi-view aggregation.
The coarse texture map is then super-resolved in UV space using the DRCT model~\cite{DRCT24} and its holes are filled using a position-aware extrapolation process, similar to the steps that lift single-view guidance to the multi-view version described in \sref{subsec: view packing}.
The final UV texture map can then be used as a hole-less, detailed, relightable mesh texture to render images with realistic view-dependent effects.

\section{Experiments}
\label{sec: experiments}

\begin{table*}[ht!]
    \centering
    \caption{Quantitative comparison between PacTure and baseline texturing methods evaluated on rendered multi-view images.
    For shaded and albedo images, we report ${\rm FID}_{\rm CLIP}$, KID ($\times 10^{-3}$), and CLIP scores.
    For roughness and metallicity images, we report RMSE.
    $\uparrow$ ($\downarrow$) means higher (lower) is better.
    We embolden the best and underline the second best results in each column.
    }
    \footnotesize
    \begin{tabular}{lccccccccr}
        \toprule
        \multirowcell{2}[-0pt]{\diagbox[height=1.7\line,width=2.5cm]{\raisebox{-5pt}{Method}}{\raisebox{0pt}{Metric}}} &
          \multicolumn{3}{c}{Shaded image} &
          \multicolumn{3}{c}{Albedo} &
          Roughness &
          Metallicity &
          \multirowcell{2.4}{Time$\downarrow$} \\
          \cmidrule(lr){2-4}
          \cmidrule(lr){5-7}
          \cmidrule(lr){8-8}
          \cmidrule(lr){9-9}
         &
          ${\rm FID}_{\rm CLIP}$$\downarrow$ &
          KID$\downarrow$ &
          CLIP$\uparrow$ &
          ${\rm FID}_{\rm CLIP}$$\downarrow$ &
          KID$\downarrow$ &
          CLIP$\uparrow$ &
          RMSE$\downarrow$ &
          RMSE$\downarrow$ \\ \midrule
        Text2Tex      & 4.533 & 1.854 & 28.89 & N/A & N/A & N/A & N/A & N/A & 1340s \\
        TEXTure       & 7.481 & 4.332 & \textbf{29.82} & N/A & N/A & N/A & N/A & N/A & 100s \\
        Paint3D        & 4.027 & 2.997 & 29.28 & 7.496 & 5.573 & \underline{29.28} & N/A & N/A & 120s \\
        FlashTex      & 6.909 & 3.575 & 29.04 & 10.15 & 7.438 & 28.29 & 0.197 & 0.482 & 1140s \\
        TexGaussian & \underline{2.868} & \underline{1.654} & 28.27 & \underline{5.265} & \underline{4.119} & 27.92 & \textbf{0.171} & \underline{0.491} & 65s \\
        PacTure (ours)  & \textbf{2.022} & \textbf{1.179} & \underline{29.40} & \textbf{3.577} & \textbf{3.820} & \textbf{29.32} & \underline{0.189} & \textbf{0.344} & \textbf{27s} \\
        \bottomrule
    \end{tabular}
    \label{tab: quantitative}
\end{table*}

We describe the dataset used in \sref{subsec: dataset}, and the training process in \sref{subsec: training}.
We compare our PacTure to baselines in \sref{subsec: comparison}, and report an ablation study in \sref{subsec: ablation}.
Further details and results are given in the appendix.

\subsection{Dataset}
\label{subsec: dataset}

To render the multi-view training set that includes maps of PBR attributes for multi-view multi-domain generation, we used the following process.

Both our training and test datasets were derived from the LVIS split of Objaverse, which consists of textured 3D meshes.
We first excluded those meshes lacking a BRDF shader or base color / metallicity / roughness attributes and removed any animation (if present) from the remaining meshes, and used the first static frame during rendering.
We also filtered out meshes with low-quality base color maps, retaining a total of 32,201 meshes with high-quality PBR materials.
Of these, 32,000 meshes were used to train our multi-view generation model, while the remaining 201 meshes formed the test set for benchmarking the entire texturing pipeline.
The mesh was normalized to $[-1,1]$ prior to rendering; we used the Blender EEVEE renderer.
We rendered the shaded image and material maps of each mesh in the training set from 6 canonical views (front, rear, left, right, top, and bottom) and used these multi-view images for training.
Meshes in the test set were instead rendered from 20 fixed views (8, 6, and 4 views with elevation 0$^\circ$, 30$^\circ$, and 45$^\circ$, respectively, and top/bottom views), and the resulting images were used as the reference to evaluate the textures generated by the methods being compared, or ablated models.
We used orthographic cameras that looked at the world origin and had the same orthographic scales.
The shaded images were rendered with a random high-dynamic-range (HDR) environment map sampled from the Laval Indoor HDR Database~\cite{LavalIndoor17} and an additional three-point lighting setup to enhance visual quality.
We rendered the PBR attributes, including metallicity and roughness, by using them as the base color and exporting the rendered base color map.
For the text prompt, we used captions from Cap3D~\cite{Cap3D23}.

\begin{figure}[t!]
    \centering
    \includegraphics[width=\columnwidth]{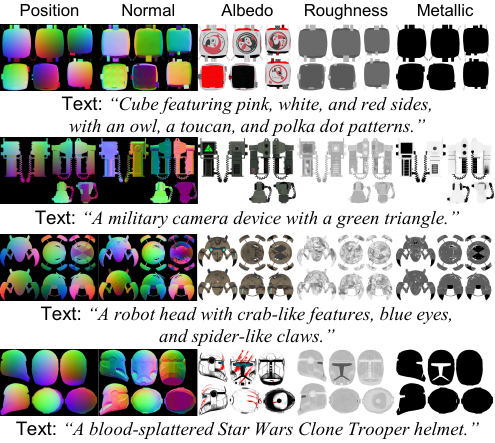}
    \caption{Each data group in our multi-view PBR dataset includes geometry conditions as model inputs and PBR material maps as training targets.}
    \label{fig: dataset example}
\end{figure}

Some examples in the resulting dataset, represented as packed multi-view maps, are shown in \cref{fig: dataset example}.

\subsection{Training}
\label{subsec: training}
To train the multi-view multi-domain generation model $\mathcal{G_{\rm MV}}$, we froze the VAE of Infinity and trained only the transformer using binary cross-entropy loss over the predicted bit token maps, using a dataset consisting of rendered shaded images and material maps for 6 canonical views (front, rear, left, right, top, and bottom) of 32,000 meshes with high-quality PBR materials, taken from the Objaverse dataset.
We employed a coarse-to-fine training strategy.
The coarse stage used a resolution of $224\times 336$ and took 72 hours on 2 NVIDIA A6000 GPUs, and the fine stage used a resolution of $832\times 1248$ and took 24 hours on 4 A6000 GPUs.

\begin{figure*}[t!]
    \centering
    \includegraphics[width=\textwidth]{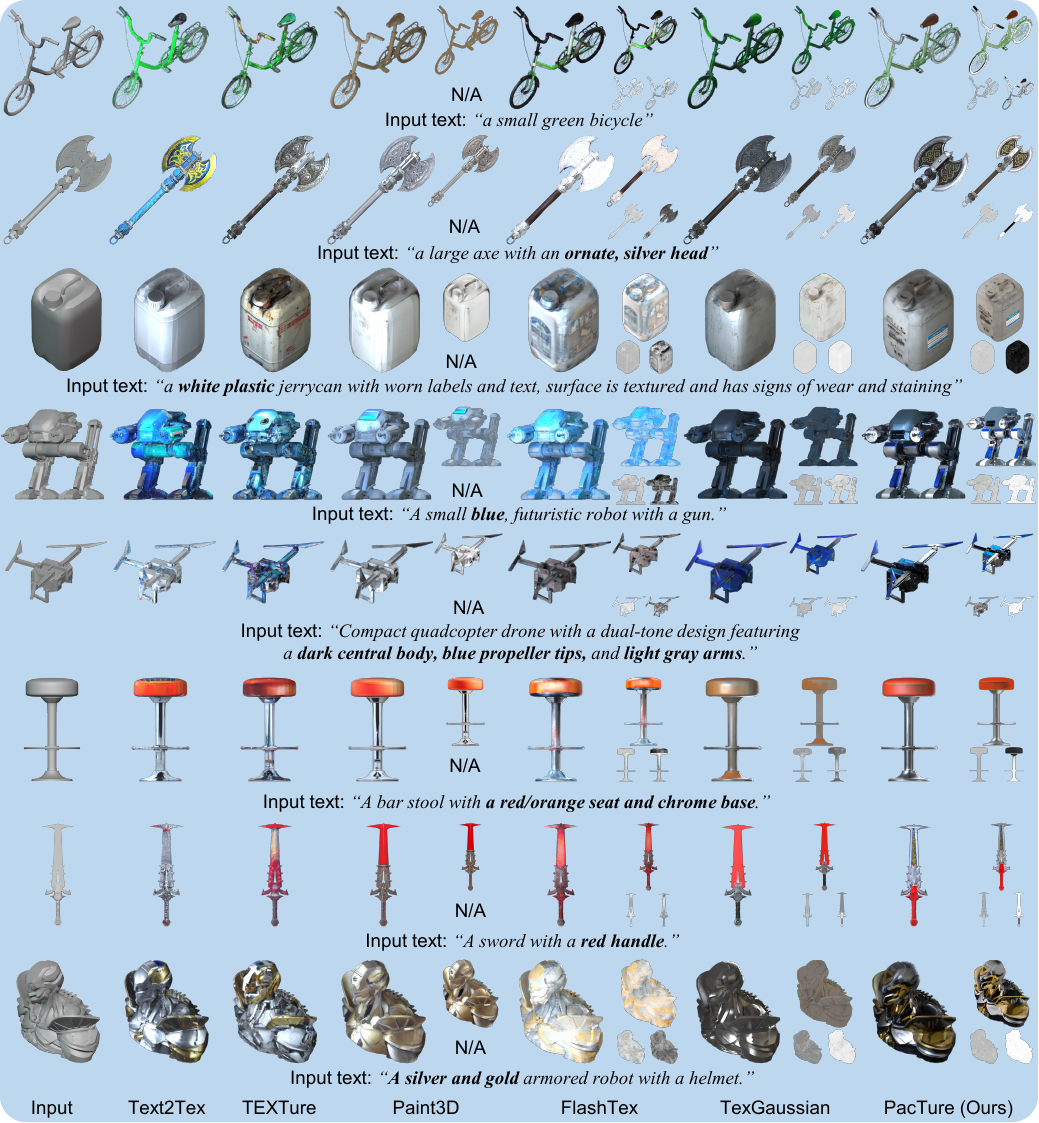}
    \caption{
    Qualitative comparison between PacTure and baseline methods.
    We show the shaded image for all methods, albedo to the right of the shaded image, and roughness/metallicity below the albedo for methods generating PBR textures.
    }
    \label{fig: comparison}
\end{figure*}

\subsection{Comparison}
\label{subsec: comparison}

\subsubsection{Comparing Methods}
As PacTure is a text-to-PBR-texture method, we compared it to five open-source text-based texture generation baselines: Text2Tex~\cite{Text2Tex23}, TEXTure~\cite{TEXTure23}, Paint3D~\cite{Paint3D24}, FlashTex~\cite{FlashTex24}, and TexGaussian~\cite{TexGaussian25}; the last two can generate PBR textures while the others only give shaded texture or albedo.
As we did not have the computational resources and data to re-implement closed-source work like Meta 3D AssetGen~\cite{Meta3DAssetGen24}, we were unable to make a comparison to these methods.

\begin{figure*}[t!]
    \centering
    \includegraphics[width=\textwidth]{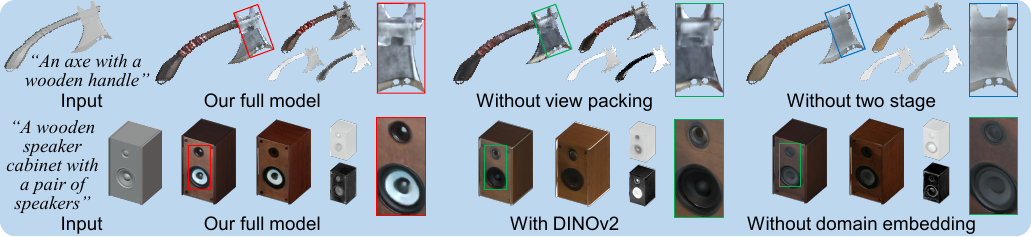}
    \caption{Ablation of various model components. The shaded image (insets showing close-ups), diffuse albedo, roughness, and metallicity images are shown for each example.}
    \label{fig: ablation}
\end{figure*}

\begin{table*}[t!]
    \centering
    \caption{User study on the generated textures. We show the one-against-one win rate for our method against each baseline, for each criterion, and the cross-participant standard deviation of win rates. A win rate $>50\%$ means that our method is favored over the competing method.
    }
    \footnotesize
    \begin{tabular}{lcccc}
        \toprule
        Method & Overall quality & Multi-view consistency & Prompt alignment & Lighting harmony \\ \midrule
        Text2Tex  & 77.9\% $\pm$ 16.2\%  & 75.8\% $\pm$ 19.1\%  & 78.3\% $\pm$ 14.3\%  & 77.1\% $\pm$ 17.4\%  \\
        TEXTure  & 82.8\% $\pm$ 19.2\%  & 82.2\% $\pm$ 19.4\% & 82.3\% $\pm$ 17.7\% & 82.8\% $\pm$ 17.9\% \\
        Paint3D  & 69.3\% $\pm$ 19.3\% & 70.5\% $\pm$ 22.4\% & 70.9\% $\pm$ 13.2\% & 69.0\% $\pm$ 17.4\% \\
        FlashTex  & 83.7\% $\pm$ 12.6\% & 82.8\% $\pm$ 12.7\% & 85.4\% $\pm$ 15.8\% & 81.8\% $\pm$ 15.2\% \\
        TexGaussian  & 74.6\% $\pm$ 14.0\% & 71.0\% $\pm$ 13.2\% & 75.5\% $\pm$ 15.0\% & 73.5\% $\pm$ 15.3\% \\
        Average  & 77.7\% $\pm$ 16.5\%  & 76.4\% $\pm$ 17.7\%  & 78.5\% $\pm$ 15.3\%  & 76.8\% $\pm$ 16.6\%
        \\ \bottomrule
    \end{tabular}
    \label{tab: user study}
\end{table*}

\subsubsection{Evaluation Protocol}
We benchmarked comparing methods using 201 meshes with PBR materials from Objaverse.
These meshes were rendered with their ground-truth textures or the textures generated by each method at $512^2$ resolution in four domains (if available): shaded, albedo, roughness, and metallicity, with the same viewpoints and lighting.
All metrics were computed using these rendered images.
For shaded and albedo images, we followed prior works~\cite{TEXGen24, Hunyuan3D2.025}, adopting the CLIP-based version~\cite{CLIP21} of FID~\cite{FID17} and KID~\cite{KID18} to evaluate texture quality and diversity.
To assess the semantic alignment of the textures with given text prompts, we employed the CLIP score~\cite{CLIP21}.
For the roughness and metallicity images, we computed the root mean squared error (RMSE).

\subsubsection{Results}
Quantitative results are summarized in \tref{tab: quantitative}.
Notably, our PacTure required only 27 s to texture a mesh on a single NVIDIA A6000 GPU, and was the most efficient.
It also outperformed all other baselines in texture quality, indicated by FID and KID.
Compared to the 3D native generation-based TexGaussian, the alignment of our generated texture with the given text prompt is better, as TexGaussian is not pretrained on a large-scale text-image dataset.
Qualitative comparisons (see \fref{fig: comparison}) further demonstrate that our method yields superior texture generation results.
Our method can generate fine details in the albedo (rows 2 and 3), and spatially-varying PBR material maps according to different parts of the object (rows 1 and 2) and the text instruction (row 3), endowing the textured object with a natural and realistic appearance under illumination, while other method generated only shaded textures or mostly spatially-uniform PBR materials.

\subsubsection{User Study}
We conducted a user study to evaluate human preference for our method over the baselines.
In each question, the participants were presented side-by-side shaded images of the textured mesh rendered by our method and a comparator method (both anonymized) from 6 views.
The participants were asked to choose the method they deemed better using each of the following four criteria:
\begin{enumerate}
\item Multi-view consistency: how consistent the object appears from different viewing angles. Good consistency means the object maintains coherent geometry, texture, and details in different views.
\item Prompt alignment: how well the generated 3D model matches the given text description. Participants were asked to consider whether the object's shape, style, features, and attributes match what was requested in the prompt.
\item Lighting harmony: the quality of lighting and shading effects. Good lighting harmony involves realistic shadows, appropriate highlights, and illumination consistent with the environment.
\item Overall quality: the overall visual quality of the generated 3D model. This includes texture detail, visual realism, absence of artifacts, and general aesthetic appeal.
\end{enumerate}
In each question, the 3D asset, the chosen 6 views, and the comparator method were all uniformly randomly sampled on-the-fly.
The results of 400 questions from 8 participants are summarized in \tref{tab: user study}.
For all baselines, our method was preferred in all four criteria.

\subsection{Ablation Study}
\label{subsec: ablation}

We conducted an ablation study on our multi-view generation model by comparing it to the following ablated variants to validate the contributions of key components:
\begin{enumerate}
\item Model without view packing, where the conditional and target multi-view grids are naively tiled instead of compactly packed.
\item Model without two-stage generation, excluding the guidance $I_{\rm MV}$ from \eref{equ: stage2}.
\item Model using a DINOv2-S encoder for the control image instead of the VAE encoder from Infinity.
\item Model without the domain embedding, in which the transformer is equipped with two output heads to generate two token maps at the same time.
\end{enumerate}
All variants were trained at a resolution of $224\times 336$ under identical settings, except for the ablated component.
\tref{tab: ablation} makes a comparison of metrics and \fref{fig: ablation} provides a visual comparison.
We observe that the proposed view packing allows more detailed texture using the same generation budget, and the incorporation of the first stage also provides richer texture.
Encoding spatially-aligned control with frozen VAE and using domain embedding for multi-domain output also improves our method.

\begin{table}[t!]
\centering
\caption{Ablation of various model components. }
\footnotesize
\begin{tabular}{lcccccccc}
\toprule
    \multirowcell{2}[-1pt]{\diagbox[height=1.8\line,width=3.6cm]{\raisebox{-2pt}{Model}}{\raisebox{2pt}{Metric}}} &
    \multicolumn{3}{c}{Shaded image} \\
    \cmidrule(lr){2-4}
    &
    ${\rm FID}_{\rm CLIP}$$\downarrow$ &
    KID$\downarrow$ &
    CLIP$\uparrow$ \\ \midrule
    Without packing   & 3.502  & 2.506  & 29.26  \\
    Without two stage & 3.340  & 2.146  & 29.05  \\
    With DINOv2     & 3.604  & 2.791  & 29.10  \\
    Without domain embedding  & 3.022  & 1.944  & 29.08  \\
    Full model    & \textbf{2.966}  & \textbf{1.753}  & \textbf{29.56}  \\
\bottomrule
\end{tabular}
\label{tab: ablation}
\end{table}

\section{Conclusions}
\label{sec: conclusion}
We have proposed PacTure, an approach for generating PBR material maps for a given mesh and text prompt, characterized by a novel view packing technique for a no-cost improvement of effective generation resolution, a two-stage single-to-multi-view pipeline for quality and flexibility, and the light-weight adaptation of next-scale prediction autoregressive models for more efficient training and inference.

Nevertheless, PacTure has some limitations.
Like other 2D-based texturing methods, PacTure cannot generate textures in regions occluded from all views.
Though we fill these regions using 3D-aware extrapolation in UV space, this may still result in unrealistic appearances upon close inspection of the occluded areas.
Also, the proposed view packing may be less effective on meshes with a cube-like 3D bounding box.
The quality of the generated multi-view texture also relies on single-view guidance, so may be reduced if the single-view generation result is poor.

\subsection*{Acknowledgements}
This work was supported by National Natural Science Foundation of China (Grant No. 62136001).

{
    \small
    \bibliographystyle{ieeenat_fullname}
    \bibliography{main}
}

\clearpage
\maketitlesupplementary

In this supplementary material, we provide additional implementation details and more experimental results of PacTure.

\section{Further Implementation Details}
\label{sec: more implementation details}

\subsection{Rendering Multi-view Geometry Conditions}
\label{subsec: pre-processing appendix}

\subsubsection{Pre-processing}
Before rendering, we normalize the positions of mesh vertices so that the center of the mesh is located at the origin and the position of every vertex is within the range $[-1, 1]^3$.

\subsubsection{Camera Setting.}
To render the conditions, we rasterize the geometric attributes of triangular faces in six canonical views using orthographic projection with identical orthographic scales for all viewpoints, such that the rendered views just fit into the viewport, leaving at least 5\% of the side length as margin.
We also record the NDC of the mesh vertices for usage in back-projection (see \sref{subsec: back-projection}).
The whole process is implemented using the python Kaolin~\cite{Kaolin19} library.

\subsection{Single-view Generation as Guidance}
\label{subsec: two stage appendix}

\subsubsection{Generative Backbones}
For the single-view generation model $\mathcal{G}_{\rm SV}$, we employ the pretrained Stable Diffusion 1.5 model~\cite{SD22}, using depth ControlNet~\cite{ControlNet23} to provide spatially-aligned geometric conditioning without further fine-tuning.
This generates a single-view shaded image, and we use Intrinsic Anything~\cite{IA24} to estimate its diffuse albedo.

\subsubsection{View Selection and Guidance Spreading}
Specifically, we select the view (from front, left, or top) that provides the largest coverage as the single view $s$ for generation.
To enhance the guidance signal and improve multi-view consistency, after obtaining the single-view albedo map, we spread the pixels in this single-view map to the pixels in other views if the latter share similar positions and surface normals with the former, using the following strategy.
As we have the geometry condition maps including position and world-space normal, we know these two attributes for each foreground pixel in the packed multi-view map.
Thus, we are able to estimate any foreground pixel's affinity to a pixel in the single view.
Specifically, we first construct a $K$-D tree~\cite{Scipy20} of pixels inside the single view using their 3D positions.
Then, for a pixel outside the single view, we query its nearest neighbor in 3D space in the $K$-D tree.
If the difference between their positions is less than 0.02 and the difference between their normals is less than 45$^\circ$, we regard these two pixels as representing the same surface location and copy the albedo value of the single view pixel to the pixel outside the single view.
After finishing this process, we use the resulting multi-view incomplete albedo map as a condition image for subsequent multi-domain multi-view generation, like other geometry condition maps.

\subsection{View Packing}
\label{subsec: view packing appendix}

\subsubsection{Patching}
As the patch size of the multi-view generation backbone, Infinity, is $p=16$, we set the size of the bin to the ceiling of atlas size divided by $p$, and the sizes of the rectangles to the ceiling of sizes of view bounding boxes divided by $p$.
In this way, the pixels of each patch come from at most one viewpoint and a single token does not contain pixels from different views.
After the optimal per-view enlargement ratio and the corresponding rotation states and packing placements have been obtained, we multiply the sizes and offsets by the patch size $p$ and put all view maps into the atlas according to this information.

\subsubsection{Rotation}
We prohibit 90$^\circ$ rotation of rectangles except for the top and bottom views, thereby preserving the visual legibility of packed images.
To allow 90$^\circ$ rotation of top and bottom views, we disallow rotation within the bin packing algorithm and rotate the rectangles outside, using a binary string to indicate the state of being rotated.

\subsubsection{Binary Search}
During each binary search over the global scaling ratio, we try every binary state of rotation during each attempt.
This is performed for at most 8 iterations.
The upper bound and lower bound of the binary search are estimated according to the total area and longest side length of view bounding boxes and the atlas.

\subsubsection{Other Details}
The paired enlargement ratio is subject to a maximum of $2\times$ magnification to avoid excessively imbalanced views or blurring.
The NDCs of mesh vertices for each view are also updated so that they now reflect the positions of vertices inside the packed multi-view maps and can be used to index pixel locations during back projection onto the UV space.

\subsection{VAR Models for Multi-view PBR Generation}
\label{subsec: multi-view generation appendix}

As mentioned in \sref{subsec: adapting VAR}, we adopt Infinity as our multi-domain multi-view generative backbone, comprising a visual (de)tokenizer and a decoder-only transformer.

\begin{figure*}[t!]
    \centering
    \includegraphics[width=\textwidth]{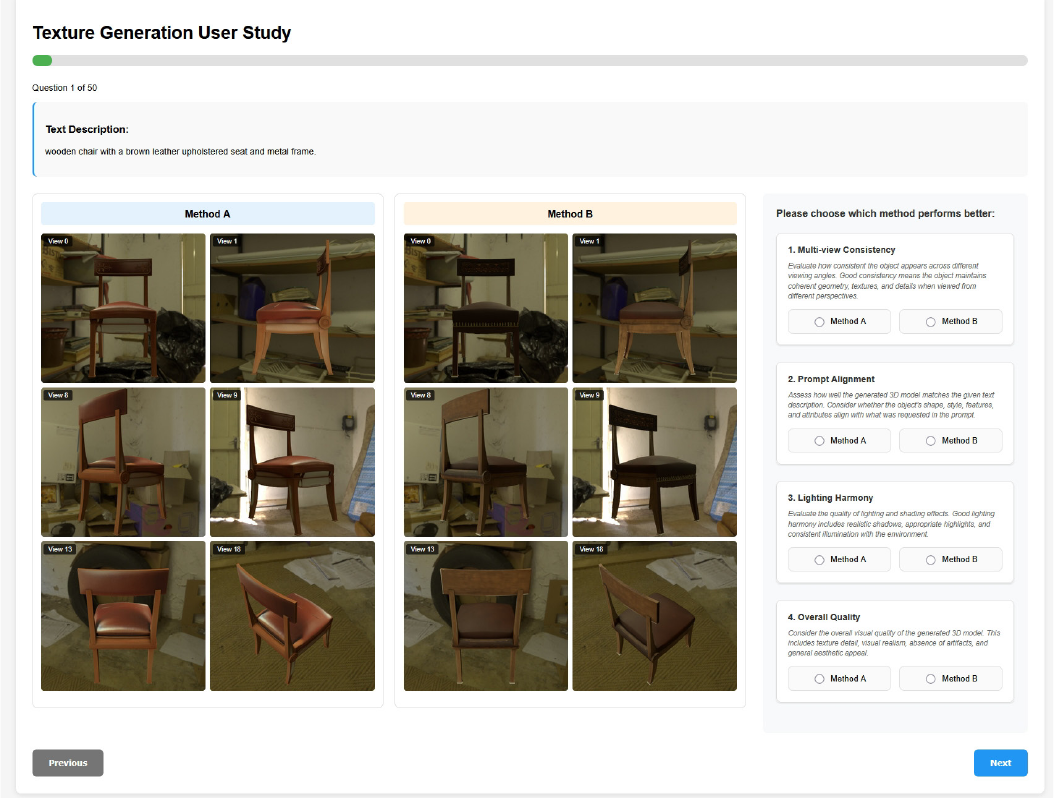}
    \caption{User interface of our user study.}
    \label{fig: user study}
\end{figure*}

\subsubsection{Network Architecture}
For the visual tokenizer, we follow VAR and Infinity, and use the published VQVAE architecture for the encoder and the decoder.
The spatial downsampling and upsampling ratio is $16\times$.
The multi-scale residual quantizer is adopted from Infinity.
The entire VQVAE is kept frozen throughout the paper.

The architecture of the transformer also follows prior works~\cite{VAR24, Infinity25}.
It uses a word embedding to project the quantized bit code (with dimensionality 32) in each place of the image token map to the same dimensionality (2048) as the encoded text embedding by Flan-T5 at the start of each scale.
Several transformer blocks (shared over all scales) then perform cross-attention between the projected image token and the encoded text embedding during text-conditioned generation.
At the end of each scale, an output head predicts the logit of each bit from the cross-attended features.
Our method injects spatially-aligned control (by encoding it using the VQVAE encoder and multi-scale quantizer) to the multi-scale residual control token maps.
At each scale, the corresponding control token map also undergoes the same word embedding as the image token map and is token-wise added to the projected image token map before the first transformer block.
The dimensionality of the domain embedding is also the same as the dimensionality of the text embedding and is added to each token before the first transformer block.

\subsubsection{Training Strategy}
We employ a coarse-to-fine training strategy: we first train with a resolution of $224\times 336$ and per-GPU batch size of 50 for 15,000 steps, taking 72 hours on 2 NVIDIA A6000 GPUs.
Then, we fine-tune the model at resolution $832\times 1248$ using a per-GPU batch size of 5 with gradient accumulation of 5 steps for another 3,000 steps, taking 24 hours on 4 A6000 GPUs.
The transformer is fine-tuned under the same setting: a base learning rate of $6\times 10^{-5}$ per 256 total batch size.
Given that we use a total batch size of $100$ (50 per GPU using 2 GPUs in the 0.06M pixel training stage, and 5 per GPU using 4 GPUs and 5 steps for gradient accumulation in the 1M pixel training stage), the real learning rate is about $2.3\times 10^{-5}$.
We use the Adam~\cite{Adam14} optimizer with $\beta_1=0.9$ and $\beta_2=0.97$.

\subsubsection{Feasibility of using VAE without Fine-Tuning.}
In our method, the VAE is used to encode the control images and decode the material maps without further fine-tuning.
We observe that, although the control images (positions, normals) differ in modality from natural images, these three modalities are reconstructed equally well by the VAE.
On our test set, the VAE reconstructs images in these domains with PSNR $\approx 25$ dB within the foreground, suggesting that directly encoding these control images via the VAE is feasible.
The VAE also reconstructs the PBR material maps (the albedo and roughness-metallicity map) well.
Thus, it is not strictly necessary to fine-tune the VAE to encode/decode images in these modalities.

\subsection{User Study}

The user interface for our user study is shown in \fref{fig: user study}; the participants are presented side-by-side rendered multi-view images of textures generated by two anonymized methods (one is ours), and are required to choose the better method for each criterion.

\section{More Results}

\begin{figure*}[t!]
    \centering
    \includegraphics[width=\textwidth]{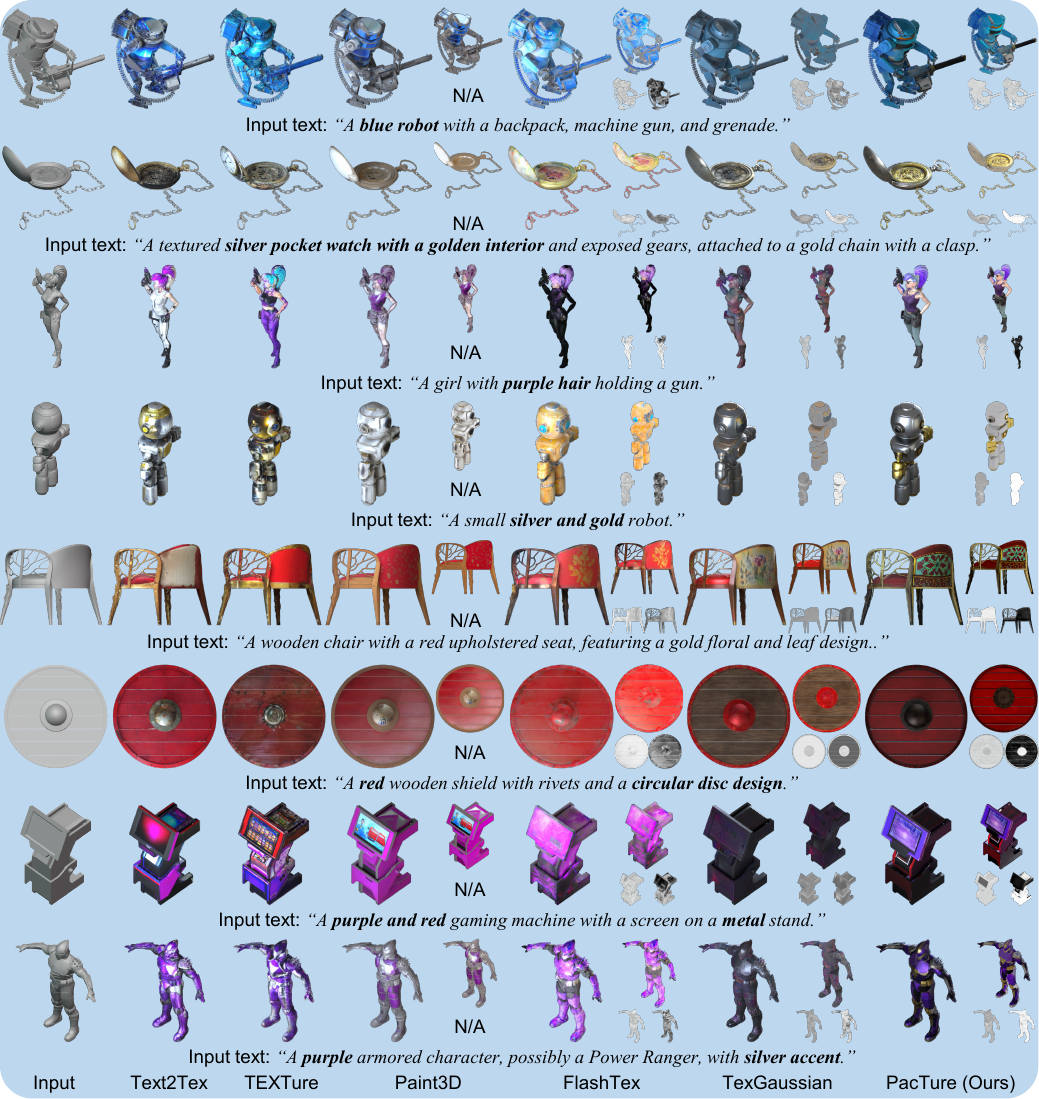}
    \caption{Qualitative comparison between PacTure and baseline methods.
    We show the shaded image for all methods, albedo to the right of the shaded image, and roughness/metallicity to the bottom of the albedo for methods generating PBR textures.}
    \label{fig: supp comparison}
\end{figure*}

In \fref{fig: supp comparison}, we provide a further visual comparison between our PacTure and other baselines.
We again observe that PacTure can generate PBR materials with fine details that follow the input prompt well.

\end{document}